\documentclass[sigconf]{acmart}

\usepackage{booktabs} 

\usepackage{algorithm}
\usepackage{algpseudocode}

\usepackage{listings}
\usepackage{placeins}

\usepackage[colorinlistoftodos,prependcaption]{todonotes}

\usepackage{color, soul}

\setcopyright{rightsretained}

\acmDOI{0.0/000_0}

\acmISBN{123-4567-24-567/08/06}

\acmConference[IWC'18]{First International Workshop on Chatbot in conjunction with ICWSM 2018}{June 2018}{Stanford, California}
\acmYear{2018}
\copyrightyear{2018}

\acmArticle{X}
\acmPrice{XX.XX}

\begin{document}
\title{Intent Generation for Goal-Oriented Dialogue Systems based on Schema.org Annotations}

\author{Umutcan \c{S}im\c{s}ek}
\authornote{Corresponding author}
\orcid{0000-0001-6459-474X}
\affiliation{%
  \institution{University of Innsbruck}
  \streetaddress{Technikerstrasse 21a}
  \city{Innsbruck}
  \country{Austria}
  \postcode{6020}
}
\email{umutcan.simsek@sti2.at}

\author{Dieter Fensel}

\affiliation{%
  \institution{University of Innsbruck}
  \streetaddress{Technikerstrasse 21a}
  \city{Innsbruck}
  \country{Austria}
  \postcode{6020}
}
\email{dieter.fensel@sti2.at}

\renewcommand{\shortauthors}{U. \c{S}im\c{s}ek and D. Fensel}

\begin{abstract}
Goal-oriented dialogue systems typically communicate with a backend (e.g. database, Web API) to complete certain tasks to reach a goal. The intents that a dialogue system can recognize are mostly included to the system by the developer statically. For an open dialogue system that can work on more than a small set of well curated data and APIs, this manual intent creation will not scalable. In this paper, we introduce a straightforward methodology for intent creation based on semantic annotation of data and services on the web. With this method, the Natural Language Understanding (NLU) module of a goal-oriented dialogue system can adapt to newly introduced APIs without requiring heavy developer involvement. We were able to extract intents and necessary slots to be filled from schema.org annotations. We were also able to create a set of initial training sentences for classifying user utterances into the generated intents.  We demonstrate our approach on the  NLU module of a state-of-the art dialogue system  development framework.
\end{abstract}

%
%



\maketitle


\section{Introduction}
Unlike more conversation oriented, human-human interaction mimicking chatbots, goal-oriented dialogue systems typically converse with humans based on defined tasks \cite{Jurafsky2017DialogChatbots}. From a natural language understanding point of view, these tasks are connected to the domain specific intents that can be identified from user utterances.
Goal-oriented dialogue systems generally work with well curated back-ends either by means of querying a database or sending requests to an API that is coupled with the dialogue system. This situation naturally makes it harder to adapt dialogue systems to different back-end systems.

After almost 30 years of its invention, the web is finally becoming more machine-oriented. This is thanks to the increasing amount of semantic annotations published on the web. The de-facto standard vocabulary for publishing semantic annotations is the schema.org vocabulary, supported by the initiative consists of Bing, Google, Yahoo! and Yandex search engines. The vocabulary contains types and properties for various domains to describe entities on the web. This semantically annotated data can be consumed by agents like dialogue systems.

A goal-oriented dialogue system should be as generic as possible to be able to operate in a heterogeneous environment like the web. In other words, it should be decoupled from the back-end. This requires not only the data but also the web services to be described semantically, so the tasks and consequently the intents that the dialogue system supports can be extracted from the web service descriptions. We consider open and flexible goal-oriented dialogue systems that can utilize different web services with the minimal human intervention is a natural step towards completing complex tasks (e.g. e-commerce) in the more machine-oriented web \cite{Simsek2018}. 

As a first step in this direction, we propose a straightforward approach for extracting intents from lightweight semantic web services described with schema.org vocabulary. In this paper we will briefly demonstrate how this vocabulary can be used to annotate Web APIs and how these annotations can be beneficial for generating intents for goal-oriented dialogue systems.

The remainder of this paper structured as follows: Section \ref{sec:RelatedWork} gives a short review of the existing efforts in this direction. Section \ref{sec:webapi-sdoactions} exemplifies the usage of schema.org vocabulary for annotating Web APIs. Section \ref{sec:generatingintents} explains our approach in detail and Section \ref{sec:usecase} demonstrates a use case with a state-of-the art dialogue system development framework. We conclude the paper in Section \ref{sec:conclusion} with a summary and identified future directions.

\section{Related Work}
\label{sec:RelatedWork}

Dialogue systems benefited from semantic technologies, especially ontologies  for a long time to represent domain knowledge in a powerful way \cite{Milward2003Ontology-BasedSystems}.  The semantic technologies are also essential elements of question-answering systems that work over linked data \cite{Unger2014}.   As mentioned in the introduction, for goal-oriented dialogues systems that go beyond question answering, semantic annotation of the data is not enough. In order to decouple a dialogue system from the Web APIs, the services should be annotated too. The functional and behavioural description of Web APIs can potentially guide generation of dialogue flows as also explained in our recent work \cite{simsek_machine_2018}. Benefiting from semantic web service descriptions for a dialogue has been explored in the literature by the SmartWeb \cite{Sonntag2007} project. They use a rule-based semantic parser \cite{Engel05robustand} to convert user utterances to a set of ontology instances. These instances are then used for querying web services described with OWL-S \cite{martin_bringing_2007}. From a natural language understanding point of view, there is no need to identify the user intent, since it is a question answering system and the only intent is to obtain some information. As for a system that can handle multiple tasks, using the backend as a driver for the natural language understanding has been proposed. A previous work demonstrated that the intents can be modelled from a backend perspective \cite{berg2011dialog}.  The work in \cite{10.1007/978-3-319-19581-0_12} adopts such a backend driven approach, but there is no semantic descriptions involved, meaning the intents that represent different tasks should be handcrafted by the dialogue system developer. 

If we consider a dialogue system as a whole, from language understanding to response generation, there has been a increasing amount of work towards using machine learning for creating  such end-to-end dialogue systems. Such systems have big advantages as they do not require any apriori knowledge, which allows them to scale well. Such systems give promising results especially for chatting oriented dialogue systems. There is also promising development regarding such dialogue systems in a more goal oriented setting \cite{bordes_learning_2017} \cite{DBLP:journals/corr/abs-1803-02279}.  

Given the developments in the machine learning for dialogue systems, approaching the NLU challenge as a classification challenge is appropriate \cite{Pappu2013}.  For industrial applications, it is very common to use dialogue system development frameworks. These frameworks also benefit from machine learning approaches for identifying the user intention based on the utterance. The classification is done with supervision, therefore the classifiers require annotated natural language statements for training. As our final goal is to generate dialogue systems based on semantically annotated data and web services,  we first start with enabling NLU modules to classify user intents with as little human intervention as possible. Increasing adoption of schema.org vocabulary gives as a strong motivation to use the vocabulary for data and web service annotations. We utilize semantic annotations in following ways: (a) to extract the intent and slots to be filled based on the high level description of a resource of a Web API (b) creating training sentences based on domain specific data stored in a knowledge graph. Although it has some limitations at the moment,  we argue the work described in the following section will be a good step towards automatically generating goal-oriented dialogue systems simply by crawling a website annotated with schema.org and schema.org actions.

\section{Web API Annotation with Schema.org}
\label{sec:webapi-sdoactions}
Schema.org actions have been added to the core schema.org vocabulary in 2014. The main idea is to be able describe not only static entities on the web, but also the actions that can be taken on them. These actions in principle may be used as a lightweight semantic web services vocabulary for describing Web APIs. We analyzed schema.org actions in the scope of lightweight semantic web services. The details of this analysis is outside of the scope of this paper and we refer the reader to our work in \cite{simsek_machine_2018}. In this section we show what such an API annotating may look like and what kind of implications does it have for generating intents for goal-oriented dialogue systems.

\begin{figure}

\includegraphics[width=0.5\textwidth]{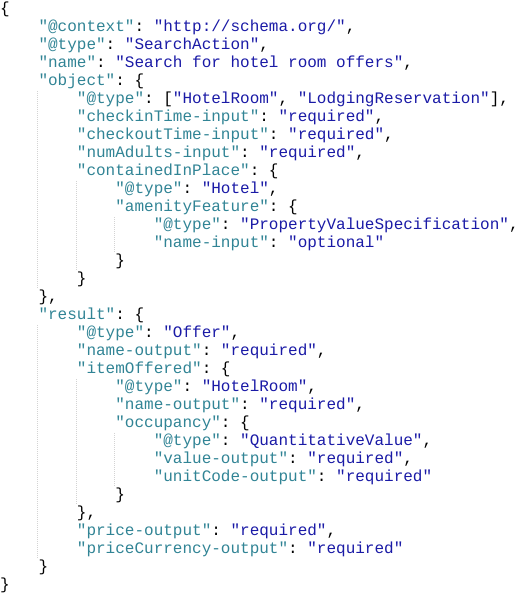}
\caption{A partial JSON-LD representation of a schema:SearchAction created to annotate Feratel API}
\label{fig:SearchAction}
\end{figure}

Figure \ref{fig:SearchAction} shows an example schema.org action annotation to describe a resource to search for hotel room offers in Feratel API\footnote{Feratel is a destination management solution provider and offers an API for e-commerce of touristic services such as accommodation.}. An action describes a resource of an API and a high level operation that can be applied to that resource. We can formalize an action $\alpha$ as a quintuple 

\begin{equation}
\alpha = (t_a, T_o, T_r, P_i, P_o)
\label{eq:action}
\end{equation}

where $t_a$ is the type of an action, $T_o$ is the set of all type values of the object of the action, $T_r$ is the set of all type values of the result of the action, and $P_i$ and $P_o$ set of input and output parameters. Figure \ref{fig:SearchAction} is missing the target property, which is not relevant for a dialogue system in terms of intent generation, but it only concerns the invocation of an action over a resource as an HTTP request. An HTTP request to the described resource may need values for the input parameters represented with \textless property name\textgreater-input properties and the API returns a set of entities as response possibly with a schema:potentialAction attached to them. These entities must contain at least the values for the properties described with the \textless property name\textgreater-output properties.

The API descriptions created with schema.org actions have interesting implications for intent generation for dialogue systems. One thing should be known that schema.org does not have a strong formal semantics \cite{Patel-Schneider_2014}. This means that the meaning of concepts is conveyed with natural language. (e.g. name and the description of the concept). We try to use such semantic information to first extract a specific intent for the operation an action represents (e.g. searching a hotel room), then by using the semantics embedded in the action annotation we will try to generate training sentences for the extracted intent. Next section explains this process in detail.

\section{Intent Generation}
\label{sec:generatingintents}
Before we go deeper in the intent generation, let us first define the term \textit{intent}. An intent is the desire to complete an action. In goal-oriented dialogue systems, user utterances may carry intents regarding a certain task. An intent contains an act, an object and modifiers for that act. The act carries a communicative function for the intent. Acts can represent generic functions such as \textit{information seeking} \cite{bunt_towards_2010} but they can also be domain specific as they are in goal-oriented dialogue systems. An object of the intent is on what the act is carried out. An act can have certain modifiers that are commonly known as frames or slots in dialogue system literature.
To explain the utterance elements with an example, let's take the following sentence: "What are the events in Innsbruck?". Here the generic act is \textit{information seeking} carried on event objects. The act is modified or filtered with an additional information, namely the location of the object. Since a goal-oriented system typically communicates with a backend, we can use a more specific act such as \textit{searching}, which is a more backend specific version of information seeking.

Currently, many state of the art NLUs use machine learning approaches for classification of intents. Therefore alongside aforementioned elements, namely act, object type and modifiers, we define one more element, which is the set of training sentences. We formalize an intent as a quadruple:

\begin{equation}
i = (a, T_o, S, M)
\label{eq:intent}
\end{equation}

where $a$ is the act, $T_o$ is the set of all object types of the intent, S is the set of training sentences and M is the set of modifiers or slots of the intent. A modifier is formalized as a triple $ m = (n, t_m, r), m \in M$, where $n$ is the name of the modifier, $t_m$ is the expected value type of the modifier and $r$ is a boolean value whether the modifier is required for the intent.

In the rest of this section we will describe how we generate intents by processing schema.org annotations. First, we describe how to collect and store the annotations in a knowledge graph to be used as the domain specific information for the generation of intents and training sentences. Then we briefly explain out straightforward methodology for generating intents. At last we show how we benefit from lexical databases and word vectors for automatically generating training sentences for the generated intents.

\subsection{Annotation Collection}

In order to reach our goal of generating intents, we collect the schema.org annotated data and web service descriptions from a website. The collection can happen automatically through a wrapper that generates schema.org annotations from a backend (e.g. relational database, Web API) or directly by crawling the website. Then we store these annotations of data and services in our knowledge graph. This way we benefit from reasoning capabilities of the graph database solution. Schema.org offers its own semantics built on RDFS. This means  some RDFS entailment rules (e.g. inheritance) can be applied to the schema.org annotations given the fact that the vocabulary is also stored in the knowledge graph.  Moreover, schema.org defines its own semantics, rather informally. For instance for inverse properties, they define an inverseOf property, whose semantics is explained with a natural language description. Such custom semantics can be introduced to the knowledge graph in terms of rules. The annotations stored here later can be used for different purposes. The action annotations are used for extracting intents and data annotations to help NLU for recognizing entities in user utterances and for training machine learning models for generating training sentences for the intents. In the next sections we will explain this processes in detail.

\subsection{Extracting Intents from Web API Descriptions}

When dealing with goal-oriented dialogue systems, the intents the dialogue system supports are directly related to the capabilities of the back-end system. If the dialogue system talks with Web APIs to complete certain tasks, then the intents are created by the dialogue system developer based on the API resources and possible operations defined on those resources. We argue that this intent generation process can be automated with the help of machine readable API descriptions, in our case with schema.org.

\begin{algorithm}
\caption{Algorithm for extracting intents}
\begin{algorithmic}
\State $A = \{a_0,a_1....,a_n\}$
\State $I = \{i_0, i_1...., i_n\}$

\ForAll {$\alpha_i \in A$}
\State i $\rightarrow a$ = getName($\alpha_i \rightarrow t_a$)
\State  i $\rightarrow T_o$ = getObjectTypes($\alpha_i \rightarrow T_o$)

\ForAll {$p_j \in \alpha_i \rightarrow P_i$}
\State add(getModifier($p_j$), $i \rightarrow M$)
\EndFor

\State add(i, I)
\EndFor

\end{algorithmic}
\label{alg:intent}
\end{algorithm}

The pseudocode in Algorithm \ref{alg:intent} summarizes the intent generation process. We iterate over the set of schema.org action annotations and extract the name of the action as well as the types of the object values. Then we create modifiers for the intent based on the input parameters of the action. The newly created intent with the act, object type and modifiers then added to the set of intents.  For example, for the action represented in Figure \ref{fig:SearchAction}, the algorithm would lead to the intent shown in Equation \ref{eq:intentSearchOffer}. 

\begin{equation}
\begin{split}
i = \{search, \{HotelRoom, LodgingReservation\},  \\
 \{\},  \\
\{(LodgingReservation.checkinTime, date, true), \\
(LodgingReservation.checkoutTime, date, true),  \\
(LodgingReservation.numAdults, date,  true),  \\
(HotelRoom.containedInPlace. \\
Hotel.amenityFeature.name, AmenityFeature ,false)\}\} 
\end{split}
\label{eq:intentSearchOffer}
\end{equation}

\begin{figure}
\includegraphics[width=0.5\textwidth]{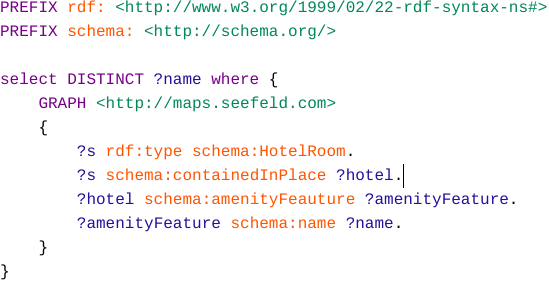}
\caption{The SPARQL query generated to retrieve hotel amenities from the knowledge graph}
\label{fig:SPARQL}
\end{figure}

The NLU modules typically need entity definitions to identify entities in the user utterances that are needed to query a database system or make an API call. These entities are usually introduced to the system by the dialogue system developer. We use semantic technologies and load the entities to the NLU based on the actions. We collect the supported entities that the NLU module needs by generating SPARQL queries to run against the knowledge graph based on the members of the set $M$. Figure \ref{fig:SPARQL} shows the SPARQL query to populate the entities that NLU module needs for recognizing the values for $HotelRoom.containedInPlace.Hotel.amenityFeature.name$ modifier. The schema.org vocabulary allows the data to be represented in several different ways. For instance, a hotel room can be connected to a hotel with schema:containedInPlace property or inversely, with schema:containsPlace property. Since the data is stored in a knowledge graph with an inference engine, the query would return all schema:amenityFeature values given that the knowledge graph contains the entailment rule for schema:inverseOf property. Moreover, the query can be optimized by querying only the named graph where the annotations for a certain website stored.

Note that at the moment the intent is generated, the set of training sentences $S$ is an empty set.  We explain the creation of training sentences for an intent in Section \ref{sec:trainingsentencegen}. 




\subsection{Automated Training Sentence Creation}
\label{sec:trainingsentencegen}

\begin{algorithm}
\caption{Algorithm for generating training sentences}
\begin{algorithmic}
 \State v = $\emptyset$ \Comment{verbs} 
 \State n = $\emptyset$ \Comment{nouns} 
 \State m = $\emptyset$ \Comment{modifiers}
 \State $I = \{i_0, i_1....i_n\}$
\ForAll {$i_j \in I$}
\State add(v, relevantWords(getSynonyms($i_j \rightarrow a$)))
\State add(n, relevantWords(getSynonyms($i_j \rightarrow T_o$)))
\State add(m, $i_j \rightarrow M$)
\State $i_j \rightarrow S$ =  buildSentences(loadCFG(), v, n, m) 
\EndFor

\end{algorithmic}
\label{alg:trainingsentence}
\end{algorithm}

The schema.org vocabulary does not have strong formal semantics\cite{Patel-Schneider_2014}, therefore to generate training sentences automatically the only thing we can rely on is the semantics hidden in the natural language descriptions of the types and properties.  For the sake of simplicity, we used a small context free grammar (CFG) to build our initial training sentences.  Algorithm \ref{alg:trainingsentence} shows our steps to generate training sentences for each intent extracted from schema.org action descriptions, based on the act, object types and modifiers. 

One challenge we encountered while creating training sentences is to put the words in a context. As we utilize the verbs and nouns appearing in the action annotations and consequently in the intents, we tried to find the relevant words. Our first attempt was to find synonyms to create varieties of training sentences of an intent. However certain words can have many different meanings as well as synonyms. This was a big issue while finding the synonyms from the WordNet lexical database \footnote{http://wordnetweb.princeton.edu}. In order to only select the context relevant synonyms, we applied the Dice coefficient \cite{dice_measures_nodate} as a string similarity measurament to description of schema.org types and the WordNet synoynms of the act ($a$) and object types ($T_o$) in order to find the relevant synoynms.  The Dice coefficient splits the strings into bigrams. As shown in Equation \ref{eq:dice}, $n_t$ is the bigrams that occur in the both strings, $n_a$ and $n_b$ represents the total number of bigrams in both strings. 

\begin{equation}
d_c = \frac{2n_t}{n_a + n_b}
\label{eq:dice}
\end{equation}

Making such a filtering helps us to avoid situations like creating a training sentence such as "I am looking for a case" for "searching for an event" intent. Alongside the synonyms, we also tried the incorporate similar verbs. For instance in the context of a goal-oriented dialogue system, "search" and "find" can be classified in the same intent from a backend perspective. Therefore we tried different similarity metrics. WordNet similarity, specifically Wu-Palmer ($wup$) metric \cite{DBLP:journals/corr/WuP94}, which uses a zero to one scale for similarity scores . Additionally, we tried to find similar words based on generic and domain specific usage. For that we used two word vector embeddings. One of them is the ConceptNet Numberbatch \cite{speer2017conceptnet}. These vectors are trained based on existing ConceptNet 5.5 knowledge graph\footnote{https://conceptnet.io} data, word2vec embeddings \cite{DBLP:journals/corr/abs-1301-3781} trained with 300B words from Google News Dataset  and GloVe embeddings \cite{pennington2014glove} trained with 840B words from CommonCrawl\footnote{http://commoncrawl.org}. The hybrid ConceptNet Numberbatch pre-trained vectors perform better than the aforementioned vectors in different evaluations \cite{speer2017conceptnet}. 
\begin{algorithm}
\caption{Algorithm for finding  relevant words}
\begin{algorithmic}
 \Procedure{relevantWords}{$synonyms$, $pos$}
 \State $rw = \emptyset$ \Comment{set of relevant words from ConceptNet Vectors}
  \State $rw_{ds} = \emptyset$ \Comment{set of relevant words from Domain Specific Vectors}
  \State $\beta = 0.5$
\State $rw$ = conceptnetVec.mostSimilar($synonyms,  pos, \beta$)
\State $rw_{ds}$ = dsVec.mostSimilar($synonyms, pos, \beta$)
\ForAll {$w_i \in rw_{ds}$}
\ForAll {$w_j \in rw$}
\If {conceptnetVec.similarity($w_i, w_j$) > $\beta$}
\State add($rw,w_i $)
\EndIf
\EndFor
\EndFor
\State
 \Return {$rw$}
 \EndProcedure

\end{algorithmic}
\label{alg:relevantwords}
\end{algorithm}

The vector embeddings consider the context of the words while learning vectors. By applying vector operations, information like word similarities, analogies can be obtained. In order to take domain specific factors into account, we trained additional vector embeddings  as Fasttext vectors \cite{bojanowski2017enriching} with descriptions of entities in the knowledge graph with the type $t \in T_o$ and its supertypes. For example, for an intent to search hotels, we trained the vectors with description of hotels and its supertypes like lodging businesses. This gave us around 2.5M words.  Typically vector embeddings work well with billions words in terms of training data, but we use this vector only as a supporting factor to include some verbs that the more generic ConceptNet Numberbatch embbeddings may be missing. Fasttext uses ngrams as the smallest units instead of words, this actually gives us the advantage of calculating similarity scores for words that do not exist in the vocabulary, which is more likely in a small vocabulary like 2.5M.  In Algorithm \ref{alg:relevantwords}, we show how similarity calculations from generic vector embeddings (conceptnetVec) and domain specific vector embeddings (dsVec) are incorporated. The \textit{relevantWords} function takes all the synonyms of a word as parameter. We first find the similar words in conceptnetVec filtered by the part-of-speech tag. Then, we compare the similar words from dsVec with the ones found in conceptnetVec and  include them into our relevant words list. If the similarity score exceeds the $\beta$ threshold. The threshold value 0.5 has been picked as an initial value after some manual observations. This threshold value can be picked more intelligently in the future.
Table \ref{table:similarity} shows some example word pairs and their similarity scores according to different calculations.  When we compare the verbs \textit{search} and \textit{find},  the vector similarity ($vec_{sim}$) gives a better score then Wu-Palmer similarity ($wup$). 

In this section, we presented a method for creating initial training sentence for extracted intents from schema.org action annotations. We utilized the act and object types of an intent. Then we enriched our word pool with synonyms and similar words we obtained from different sources. This allows us to build not only sentences like "search a hotel room", but also "find a hotel room", even though search and find are not synonyms but they have a certain semantic relationship.  In the next section we will show how the generated training sentences and intents look when loaded to a state-of-the-art NLU . 

\begin{table}
    \begin{tabular}{|l|l|l|}
    \hline
    \textbf{Word Pair}   & $vec_{sim}$ & $wup$  \\ \hline
    search-find & 0.67    & 0.33 \\ \hline
    search-need & 0.51    & 0.33 \\ \hline
    search-look\_around & 0.57    & 0.33 \\ \hline
    \end{tabular}
    \caption{Similarity score comparisons between vector embeddings and WordNet (Wu-Palmer)}
    \label{table:similarity}
\end{table}

\section{Use Case: Generating Intents for DialogFlow}
\label{sec:usecase}

Since the last couple of years, the frameworks that enable dialogue system development gained popularity. There are now many examples of such frameworks like DialogFlow\footnote{http://dialogflow.com}, Wit.ai\footnote{http://wit.ai} and relatively new Amazon Lex\footnote{https://aws.amazon.com/lex}. The advantange of these frameworks is that they provide a lot of out-of-the-box features to ease many aspects of dialogue system development such as natural language understanding and dialogue management. They also offer mechanisms to integrate the bots with several applications like Facebook Messenger or Slack, which helped to gain a vast industrial adoption in a rather short period of time. Therefore in this section, we decided to demonstrate our approach on the NLU modules of one of these frameworks, namely DialogFlow. The other frameworks in the same direction also operate in a similar way, therefore our implementation should work with other frameworks as well as custom developed systems with similar principles with minimal effort.

In Figure \ref{fig:SearchAction}, we showed a schema:SearchAction created for a real hotel booking API. Based on that action, we extracted the intent as shown in Equation \ref{eq:intentSearchOffer}. In this section, we will present how does the extracted intent look in DialogFlow. In DialogFlow, a basic intent has a name, training sentences and slots that are needed to be filled to fullfil the intent. Given the intent $i$, we choose a naming scheme such as $a.t_0,-t_1...t_n$, where $t \in T_o$. That means, for the intent in Equation \ref{eq:intentSearchOffer}, \textit{search.HotelRoom-LodgingReservation} is generated as the intent name. The slots are also generated based on  $m \in M$ extracted from schema:SearchAction. For example, in the case of the intent in Equation \ref{eq:intentSearchOffer}, the name of the slot for check-in time is generated as   \textit{object.LodgingReservation.checkinTime}, which additionally indicates the JSON path of the input property, which is useful for the backend logic that fulfills the intent. As for the training sentences, the Algorithm \ref{alg:trainingsentence} uses the act $a$ and name of the object types $t \in T_o$ as well as a sample of possible values for each modifier $m$ from the knowledge-graph. If value cannot be found in the knowledge graph, then a placeholder is created instead. For the intent in Equation \ref{eq:intentSearchOffer} a list of possible training sentences is shown in Listing \ref{lst:trainingsentences}.

\begin{minipage}{0.9\linewidth}
\begin{lstlisting}[numbers=left, stepnumber=1, xleftmargin=2.5em, framexleftmargin=2em, frame=single, columns=fullflexible, caption=A sample of generated training sentences, captionpos=b, label=lst:trainingsentences]
I search for a hotel room
We look for a hotel room
search for a hotel room,
We search for a lodging reservation
We want to search a hotel room
find a hotel room with free wifi
find a hotel room with wellness
find a hotel room on 27.05.2018
looking for an accommodation
I want to discover a lodging reservation 
rout up a hotel room
We are seeking  a hotel room
I explore a lodging reservation
I am looking around for a hotel room for 4
I want to search for a hotel room from 27.05.2018 to 30.05.2018
I want to look for a hotel room
We are looking  a hotel room
We want to explore a hotel room
\end{lstlisting}
 \end{minipage}
 
The sentences are built automatically based on the results of the algorithms described in Section \ref{sec:trainingsentencegen}. The small context free grammar is used for creating possible grammatical variations for the generation. At the moment, the prepositions required for inserting the modifiers are statically specified in the grammar (e.g. Lines 5-8). While the verbs \textit{"search, look for, explore"} are retrieved from WordNet, \textit{find, rout up, look around"} are retrieved from the vector embeddings. Similarly, word \text{accommodation} is retrieved from the vector embeddings as it is related to the phrase  \textit{"hotel room"}. There are some generated sentence that do not make too much sense, such as "I want to discover a lodging reservation", but these do not make too much of a difference while classifying user utterances by intents.

\begin{figure*}[!htb]
\frame{\includegraphics[width=\textwidth]{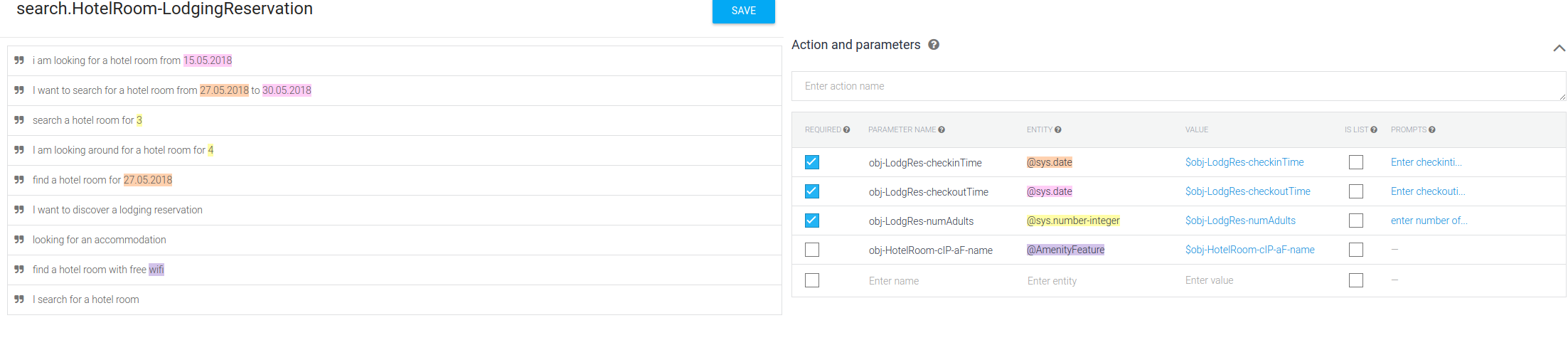}}
\caption{Generated intents, slots and training sentences loaded to DialogFlow}
\label{fig:dialogflow}
\end{figure*}

Figure  \ref{fig:dialogflow} shows the generated intent and training sentences added to DialogFlow. For implementation specific reasons, we created a mapping of certain data types from schema.org (e.g. schema:Number, schema:Date) to built-in datatypes of DialogFlow. The intents are loaded to DialogFlow via the API programmatically.

In this section, we demonstrated our approach with a state-of-the-art dialogue system development framework. The NLU module requires annotated training sentences for user utterance classification, therefore we provided automatically generated natural language sentences to train the machine learning model in the background.

\section{Conclusion}
\label{sec:conclusion}

As the web evolves to be a more machine-readable platform, the task of consumption of the content uncovers new challenges. Goal-oriented dialogue systems are prominent agents to consume this machine readable content. However, creating a dialogue system that can adapt to new data and services is not trivial, since every web service requires different workflows for completing certain tasks. This reflects to dialogue systems in terms of dialogues, which means that the dialogue systems should be able to guide the user with an appropriate dialogue according to the web service workflow. Towards this direction, we argue that the first step is that a dialogue system understands what user wants to do. To that end, we developed and demonstrated an approach for generating intents based on the lightweight semantic web service descriptions with schema.org vocabulary. To achieve our goal, we also utilized a knowledge graph to benefit from the domain specific knowledge.
Given the fact that many state-of-the-art dialogue system development frameworks use supervised machine learning for NLU, we also introduced a method for generating training sentences for the intents. At the moment, we can generate variety of sentences by utilizing lexical databases and word vector embeddings, however our method still has its limitations. The main limitation is the generation of meaningful sentences with modifier values. We see this problem as a text imputation or sentence completion problem and currently investigating solutions with LSTM neural networks for the future work, as well as better ways to incorporate sparse domain specific text for training the ML models. Another limitation is the selection of threshold value $\beta$ for word similarity. We plan to find a better way to find this threshold value based on the statistical analysis of the data.

We see the work explained in this paper as a contribution also towards automated response generation, which is one of the next steps for automated generation of dialogue systems based on web service descriptions. Although we showed some results of this approach, the real evaluation would be to see (a) if the generated sentences make sense linguistically (b) how a dialogue system with automatically generated intents perform in comparison to a dialogue system with manually created intents. In the future work, we will address the aforementioned limitations and evaluate the dialogue system with automatically generated intents from natural understanding point of view\footnote{The demo bot can be found here: https://bot.dialogflow.com/3aa58719-b665-4e7b-970a-564c1b9a64c5} with experiments involving real users in real use cases like e-tourism.

\section*{Acknowledgement}
The authors would like to thank the members of the semantify.it team (Richard Dvorsky, Thibault Gerrier, Roland Gritzer, Philipp H\"{a}usle, Omar Holzknecht, Elias K\"{a}rle and Dennis Sommer) for fruitful discussions and implementation support. 

\FloatBarrier

\bibliographystyle{ACM-Reference-Format}
\bibliography{references.bib}

\end{document}